\pdfoutput=1

\documentclass[11pt]{article}

\usepackage{EMNLP2023}

\usepackage{times}
\usepackage{latexsym}
\usepackage{graphicx}
\usepackage{amsmath,amsfonts,amssymb}
\usepackage{mathtools}
\usepackage{caption}
\usepackage{lipsum}
\usepackage{booktabs}
\usepackage{multirow}
\usepackage{soul}
\usepackage{xcolor}
\usepackage{arydshln}
\colorlet{myPurple}{blue!60!red}
\usepackage{color, colortbl}
\definecolor{Gray}{gray}{0.9}
\definecolor{ao(english)}{rgb}{0.0, 0.5, 0.0}
\definecolor{cardinal}{rgb}{0.77, 0.12, 0.23}
\usepackage{url}
\usepackage{float}
\usepackage{tabularx}
\usepackage{appendix}
\usepackage{placeins}
\usepackage{enumitem}
\usepackage{array}
\setlist[itemize]{leftmargin=*}
\makeatletter
\newcommand{\ssymbol}[1]{^{\@fnsymbol{#1}}}
\makeatother
\newcolumntype{?}{!{\vrule width 3pt}}
\usepackage{pdflscape}
\usepackage{rotating}

\usepackage[T1]{fontenc}

\usepackage[utf8]{inputenc}

\usepackage{microtype}

\usepackage{inconsolata}

\title{Tracing Influence at Scale: A Contrastive Learning Approach to Linking Public Comments and Regulator Responses}

\author{First Author \\
  Affiliation / Address line 1 \\
  Affiliation / Address line 2 \\
  Affiliation / Address line 3 \\
  \texttt{email@domain} \\\And
  Second Author \\
  Affiliation / Address line 1 \\
  Affiliation / Address line 2 \\
  Affiliation / Address line 3 \\
  \texttt{email@domain} \\}

\author{Linzi Xing$^\dagger$, Brad Hackinen$^\ddagger$, and Giuseppe Carenini$^\dagger$\\
  $\dagger$ University of British Columbia, Vancouver, Canada \\
  $\ddagger$ Ivey Business School, London, Canada \\
  \tt \{lzxing, carenini\}@cs.ubc.ca \\
  \tt bhackinen@ivey.ca 
  }

\begin{document}
\maketitle
\begin{abstract}
U.S. Federal Regulators receive over one million comment letters each year from businesses, interest groups, and members of the public, all advocating for changes to proposed regulations. These comments are believed to have wide-ranging impacts on public policy. 
However, measuring the impact of specific comments is challenging because regulators are required to respond to comments but they do not have to specify which comments they are addressing.
In this paper, we propose a simple yet effective solution\footnote{\url{https://github.com/bradhackinen/comment_response_linking}} to this problem by using an iterative contrastive method to train a neural model aiming for matching text from public comments to responses written by regulators. We demonstrate that our proposal substantially outperforms a set of selected text-matching baselines on a human-annotated test set. 
Furthermore, it delivers performance comparable to the most advanced gigantic language model (i.e., GPT-4), and is more cost-effective when handling comments and regulator responses matching in larger scale.


\end{abstract}

\section{Introduction}
Policymakers rely on information provided by external stakeholders to help design new regulations. For U.S. federal regulators, this process is formalized by the Administrative Procedures Act which requires that whenever an agency is going to make a policy change (known as a "rule"), they must first publish a proposed rule and accept public comment. Then, in the final rule, the agency must respond to comments they received. 
The number of comments received by regulators has been growing over time, and the federal government now regularly received more than a million comments per year.

Existing research suggest that public comments can have substantial impacts on public policy \cite{doi:10.1146/annurev-polisci-050817-092302}. However, measuring the influence of individual organizations or tracking patterns of influence over time has been limited by the challenging nature of the data. Both comments and regulator responses are in gigantic scale and take the form of complex natural language text. Prior attempts at large-scale analysis have borrowed insights from the research field of NLP by measuring the lexical overlap between comments and rule text
, with researchers assuming that a high degree of overlap is suggestive of influence \cite{10.1093/qje/qjab023, dwidar_2022, carpenter2022inequality} . However, this approach provides at best a noisy measure of influence, which is difficult to verify. Therefore, we aim for pursuing a more precise and efficient measure based on analyzing the regulator's responses to comments and then matching comments to specific responses. Given that some responses are positive, with agencies accepting commenter's suggestions while others are negative, with the agency rejecting the comment, it is very important to link the right comments to the right responses.

In this paper, we propose a simple yet effective iterative contrastive learning paradigm to train a neural-based comment-response matcher in an unsupervised manner. Specifically, we first construct a pseudo training dataset comprising of hard positive and negative samples generated by the inital setup of our proposed comment-response matcher (SBERT \cite{reimers-gurevych-2019-sentence} as the backbone). This matcher is then optimized on the obtained training pseudo data and subsequently utilized to generate the hard positive and negative examples for the next iteration. Through empirical evaluation on a human-annotated test set, our proposed comment-response matcher not only surpasses selected unsupervised text-matching benchmarks utilized in previous literature but also achieves comparable performance with the state-of-the-art gigantic language model -- GPT-4 \cite{openai2023gpt4}, while remaining more cost-effective to deploy on the full-scale comment-response matching.

\section{Comment-Response Matcher}
In this paper, we aim to design a text matching model (Section~\ref{sec:mr}) that can effectively and efficiently assess the semantic relevance between the public comment text and responses produced by regulators.
In essence, given a comment chunk from public $c = \{c_1, \cdots , c_m\}$ and a regulator's response $r = \{r_1, \cdots, r_n\}$, where each $c_k$ is a token in the comment and each $r_k$ is a token in the response, our goal is to learn a function $f : (c, r) \rightarrow s$ that predicts the score $s$ indicating the likelihood that comment $c$ and regulator's response $r$ pertain to the same topic, and that the concern in $c$ is addressed in $r$.

\begin{figure}
    \centering
    \includegraphics[width=3.05in]{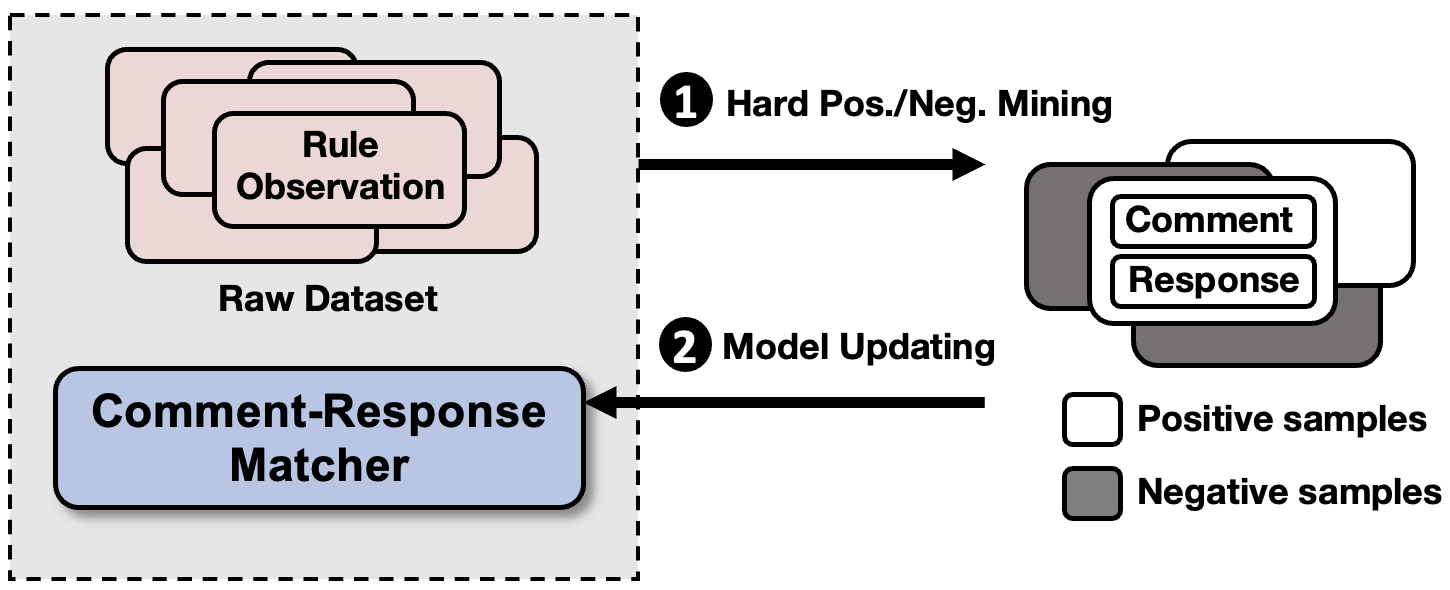}
    \caption{An overview of the iterative training scheme for our proposed comment-response matcher. }
    \label{fig:training_scheme}
\end{figure}

As illustrated in Figure~\ref{fig:training_scheme}, we employ an iterative contrastive learning paradigm with the training procedure (Section~\ref{sec:train}) 
consisting of two steps performed alternatingly, namely \textit{hard pos./neg. mining} and \textit{model updating}.

\subsection{Model Architecture}
\label{sec:mr}
Our proposed comment-response matcher functions as a binary classifier essentially comprising two components: a \textbf{text encoder} with \texttt{SBERT}\footnote{We also considered BERT \cite{devlin-etal-2019-bert}, RoBERTa \cite{DBLP:journals/corr/abs-1907-11692}, Legal-BERT \cite{chalkidis-etal-2020-legal} and other different versions of SBERT for text encoder, but eventually chose SBERT with version of \textit{multi-qa-mpnet-base-dot-v1} for its observed superior performance.} \cite{reimers-gurevych-2019-sentence} as its underlying structure, followed by a \textbf{scoring layer} yielding the likelihood of a pair of comment and response being a match.
More formally, given a pair of randomly sampled comment chunk and response $c_i$ and $r_j$, we first separately acquire the embeddings for these two textual units:
\begin{equation}
    {v}_{c_i} =  S\text{-}BERT(c_i), \label{eq:1}
\end{equation}
\vspace{-4ex}
\begin{equation}
    {v}_{r_j} =  S\text{-}BERT(r_j)\space \label{eq:2}
\end{equation}
Then the probability of $r_j$ responds to $c_i$ (a match) is computed with the negative exponent of cosine distance between ${v}_{c_i}$ and ${v}_{r_j}$:
\begin{equation}
    p(match|{v}_{c_i},{v}_{r_j}) = exp(-\alpha*(1-{v}_{c_i}\cdot{v}_{r_j})) \label{eq:3}
\end{equation}
where $\alpha$ serves as a hyper-parameter that controls the decay rate of the matching probability. A greater value of $\alpha$ results in a more pronounced decrease in matching probability when cosine distance increases.
Throughout the training process, we optimize the model with cross-entropy loss.

\subsection{Training Scheme}
\label{sec:train}
Generally, we use contrastive learning paradigm \cite{1640964} to train our proposed comment-response matcher. More concretely, we optimize the text encoder in the matcher on selected hard positive and negative samples to effectively capture signals indicating the semantic relevance between public comments and responses from the regulator. This process is therefore arguably conducive to accurately predict whether a comment is discussed in a given response.
The training scheme for the matching model spans several iterations, with each iteration consisting of two steps: \vspace{0.8ex}
\\
\textbf{-1: Hard Pos./Neg. Mining.}
As illustrated in Figure~\ref{fig:training_scheme}, our preliminary regulatory data for rulemaking is structured in the form of \textbf{rule observations} (\S\ref{sec:exp}). Each of these rule observations consists of a set of comment chunks and a set of responses associated with a particular rule. As this raw dataset does not have any explicit ground-truth labels about matching between responses and comments within the rule, 
we make the model training entirely rely on the labels of created pseudo positive and negative comment-responese sample pairs.
To do so, we first identify a set of "positive pairs" from the raw data.
More specifically, for each response, we find its most similar comment chunk within the same rule observation. This similarity is calculated based on the embeddings of the model's text encoder optimized from the prior iteration. In this way we obtain 11,828 positive comment-response pairs. 

In order to improve the robustness and efficiency of model training, within one training step, we first draw a batch of $M$ comment/response strings and then extract hard positive and negative samples associated with strings in the batch. Subsequently, we update the encoder-based matching model on these hard positive/negative samples utilizing in-batch contrastive learning \cite{wu-etal-2020-tod, zhou-etal-2022-learning}.
In practice, we initially apply the matching model, derived from the last training iteration, to all comment/response strings, yielding a total of $11,828\times2=23,656$ embeddings. We then pair each of the $M$ strings in the sampled batch with all embeddings, compute the loss, and generate a loss matrix $l \in \mathbb{R}^{M\times 23656}$.
Subsequently, we perform \textit{argmax} on each row of $l$ to identify the response-comment pair corresponding to the maximum loss, ultimately producing $M$ hard positive/negative samples. Each hard positive sample refers to a possibly matched pair which the model struggles to allocate high matching probability to, whereas each hard negative sample refers to a possibly mismatched pair to which the model tends to assign high matching probability.
\vspace{0.8ex} \\
\textbf{-2: Model Updating.}
Once hard positives/negatives for a training step are obtained, in this phase, we update weights of the comment-response matching model by minimizing the cross-entropy loss as described in Section~\ref{sec:mr}. This allows us to pull the matched comments and responses closer and push the unmatched ones far apart. The model updated in the current iteration will be fixed and serve as the text encoder to mine hard positive and negative samples again for the next training iteration.

\section{Experiments and Analysis}
\begin{figure*}
\centering
\includegraphics[width=6.2in]{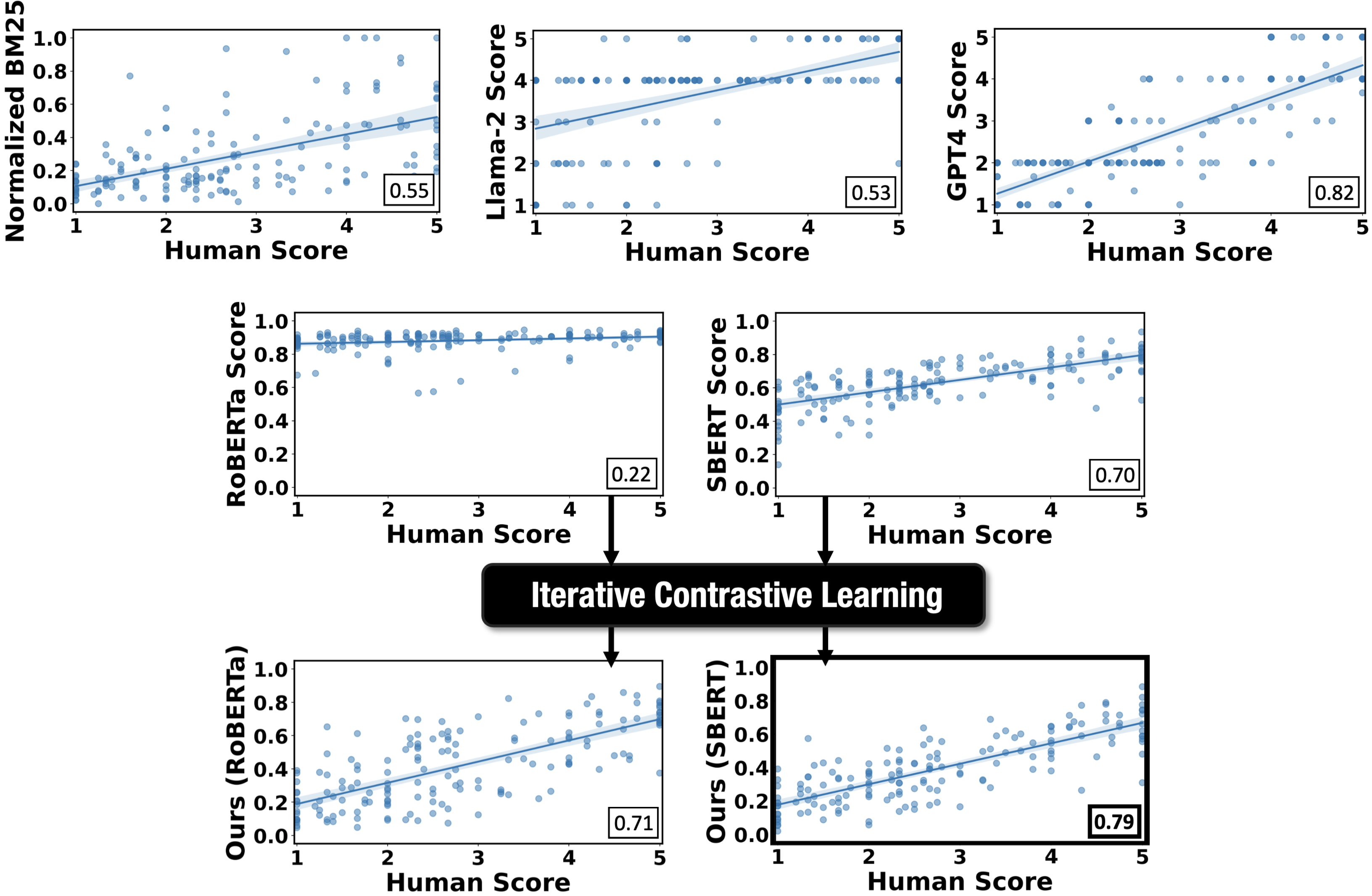}
\caption{\label{fig:correlations}  Scatter plots illustrating the correlation between human judgement and seven comment-response matching methods (including Ours (RoBERTa) and Ours (SBERT), which are RoBERTa and SBERT applied our iterative contrastive learning framework) on the 160 test samples. The Pearson’s correlations are shown at bottom-right. The best performance achieved by our proposal is highlighted in the bolded box.}
\end{figure*}

\subsection{Experimental Setup}
\label{sec:exp}
\textbf{Datasets.} 
As mentioned in \S\ref{sec:train}, our preliminary regulatory data for rule-making is structured in the form of rule observations where each rule observation is with a hierarchy depicted as follows:

\begin{itemize}
    \item \textbf{A rule observation} (about one rule document)
    \begin{itemize}
        \item \textbf{A set of comment documents} associated with the rule (Comment A, B, $\ldots$)
        \begin{itemize}
            \item \textbf{A set of comment chunks} in Comment A (Comment A-1, A-2, $\ldots$)
            \item \textbf{A set of comment chunks} in Comment B (Comment B-1, B-2, $\ldots$)\\
            $\vdots$
        \end{itemize}
        \item \textbf{A set of regulator’s responses} extracted from the rule document (Response A, B, $\ldots$)
    \end{itemize}
\end{itemize}

\noindent Textual data in rule observations comes from two main resource: rules published in the Federal Register from 2000-2022, and comments submitted to \textit{regulations.gov} from 2000-2022. As one rule document contains paragraphs other than regulator responses to public comments (e.g., background information, summary of comments), we extract only responses
from each rule document using a supervised paragraph classifier developed for another parallel research project.
We leverage some external metadata of rules publicly accessible in \textit{federalregister.gov} and \textit{reginfo.gov} to attach comments to the rules downloaded from the different resource.
As one comment document can be extensively long, we chop it into a series of comment chunks by grouping adjacent paragraphs in the comment following 1000 token limit. Paragraphs longer than 1000 tokens are deemed as single chunks.
After applying some pre-processing constraints (more details in Appendix~\ref{app:data_cosntruction}), we finally obtain a dataset covers 6,727 rules, 17,452 responses, 10,456 comments chopped into 193,143 comment chunks.

For test data construction, we uniformly (see Appendix~\ref{app:data_cosntruction}) sample 160 pairs of comment chunk and response from all possible combination in the dataset, and recruite seven students from the law program of our institution to annotate this test set. Annotators were asked to score the relevance of each comment chunk to the accompanying response using a 5-point Likert scale (see Appendix~\ref{app:prompt_templates} for detailed annotation instructions). Each sample was assigned to multiple annotators, thus we received 3-5 independent evaluations for each testing pair.

We include more details about our dataset construction pipeline in Appendix~\ref{app:data_cosntruction}.
\vspace{0.8ex} \\
\textbf{Baselines.}
We compare our proposal with three baseline text matching algorithms (see Figure~\ref{fig:correlations}). They are: (1) \textbf{Normalized BM25} \cite{10.1561/1500000019}, as a widely used term weighting-based ranking model usually applied for information retrieval. We calculate BM25 scores for the corresponding responses and comments tied to the same rule. These scores are then normalized on a per-rule basis; 
(2) \textbf{RoBERTa Score} \cite{DBLP:journals/corr/abs-1907-11692}, which employs the vanilla RoBERTa\textsubscript{base} as text encoder, transforming both comments chunks and responses into embeddings, which are then used for matching score computing. As we employ the same scoring layer (in Section~\ref{sec:mr}), this baseline is essentially equivalent to our proposed matching model in iteration 0; 
(3) \textbf{SBERT Score} \cite{reimers-gurevych-2019-sentence}, which employs the SBERT (multi-qa-mpnet-base-dot-v1)\footnote{\url{https://huggingface.co/sentence-transformers/multi-qa-mpnet-base-dot-v1}} as text encoder. The score computing of this baseline is in a manner similar to the RoBERTa Score introduced above;
(4) \textbf{Llama-2-Chat (70B)} \cite{touvron2023llama}, currently the top-performing fundamental gigantic language model within the open-sourced Llama family. We essentially deem it as a human evaluator by providing it with the same guidelines giving to human annotators and then task it to assign a score on the 5-point Likert scale for each pair of comment and response;
(5) \textbf{GPT-4} \cite{openai2023gpt4}, currently the state-of-the-art gigantic language model, leading in both open-sourced and closed-sourced domains. We prompt it to assign scores for comment-response pairs in the same manner as Llama-2-Chat (70B) introduced above\footnote{The detailed prompt for Llama-2-Chat (70B) and GPT-4 is in Appendix~\ref{app:prompt_templates}.}.

\vspace{0.6ex}
\noindent
\textbf{Implementation Details.}
As in Section~\ref{sec:mr}, we use \texttt{SBERT}\texttt{(multi-qa-mpnet-base)} \cite{reimers-gurevych-2019-sentence} as the backbone text encoder to demonstrate our proposed comment-response matcher, given its superior performance. However, to validate the model-agnostic nature of our proposed iterative contrastive learning framework, we also test with the vanilla \texttt{RoBERTa\textsubscript{base}} \cite{DBLP:journals/corr/abs-1907-11692} as an alternate backbone text encoder, aiming to discern if improvements brought by the iterative contrastive learning framework extend beyond just one particular text encoder. For both, We take the mean of the contextualized representation of the last hidden layer as text embeddings. For the scoring layer, we set hyper-parameter $\alpha = 50$. For training, we use AdamW \cite{Loshchilov2017DecoupledWD} with with $lr = 1e^{-5}$ and batch size $= 8$. We conduct 5 iterations of model training, with each iteration detailed in \S\ref{sec:train}.

\subsection{Experimental Results}
To investigate how well the baselines and our proposed comment-response matching model align with human judgments, in Figure~\ref{fig:correlations}, we use scatter plots to visualize their correlations with human scores, as well as report the Pearson’s $r$ correlation score.
We can observe that even though GPT-4's predictions show the highest correlation with the 5-point Likert human annotations, our proposed matching model also demonstrates strong performance as ranked in the second place, outperforming all left baselines 
by a considerable margin.

More concretely, 
BM25 tends to underestimate the relevance between comments and responses, assigning low scores to many pairs that humans consider highly relevant in topic.
As sharing the same rating scale with human, the predictions of GPT-4 align closely with human judgements, whereas Llama-2-Chat (70B) correlation with human is way less desirable. Interestingly, GPT-4 demonstrates a strong tendency to consistently assign score `2' to samples that humans rated within the range of [1,3], which may indicate that GPT-4 is cautious to determine a comment-response pair as entirely irrelevant.
The vanilla RoBERTa without any tuning on our dataset extremely overestimates the relevance between comments and responses by assigning high similarity scores indiscriminately to both matched and unmatched sample pairs.
On the other hand, SBERT, being a superior text matching model pre-trained on semantic search as a close analogue to our task, aligns more closely with human judgment, yet the similarity scores it produces for both matched and unmatched samples still fall within a relatively narrow range.
When our proposed contrastive learning framework is applied to RoBERTa and SBERT, the correlation of these two base text encoders with human judgments increases from 0.22 and 0.70 to 0.71 and 0.79 respectively, bringing the improved SBERT's performance remarkably close to that of GPT-4 (0.82).
It demonstrates the model-agnostic behavior of our iterative contrastive learning framework when effectively interacting with different base encoders. Hence, we believe that with a more advanced base encoder, we could potentially match or even surpass the performance of GPT-4.

\begin{figure}
\centering
\includegraphics[width=2.98in]{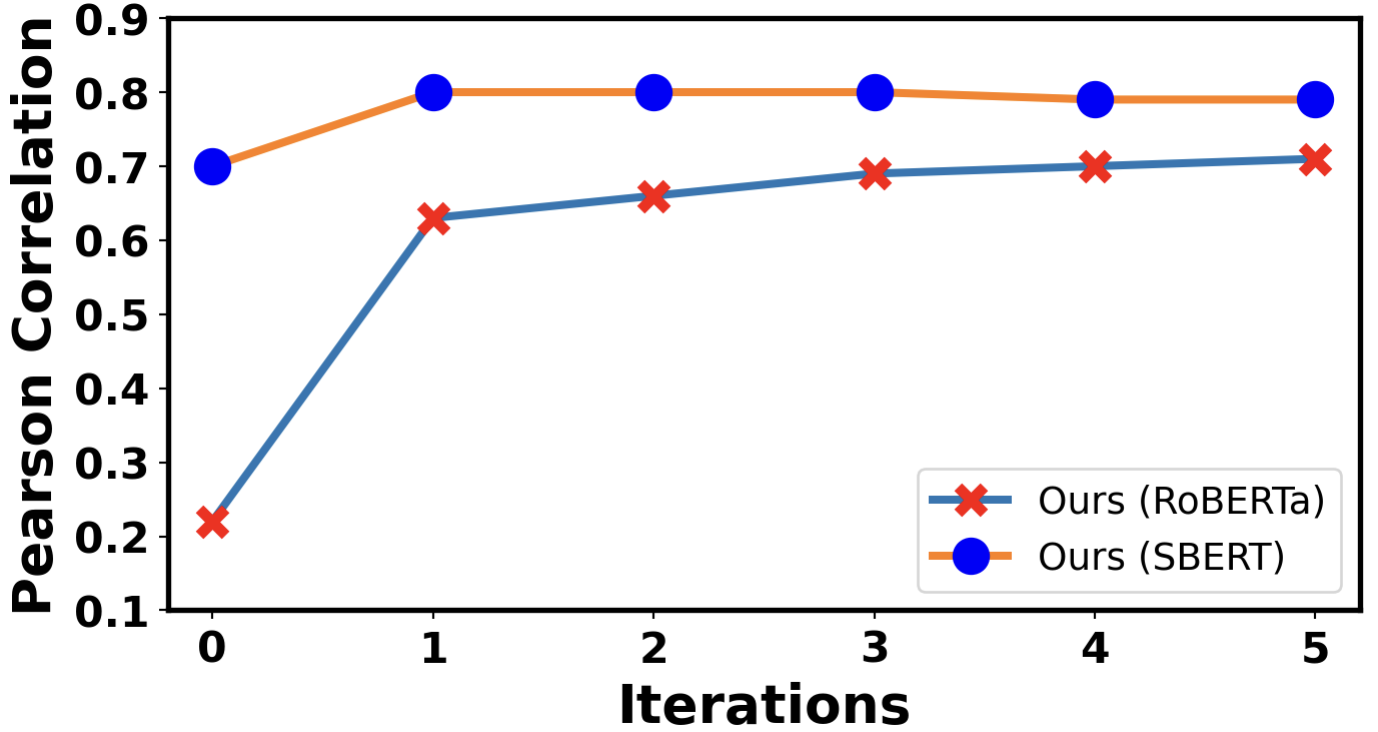}
\caption{\label{fig:curve} The performance (Pearson's correlation) of the RoBERTa- and SBERT-based comment-response matcher on our test set after each training iteration.
“Iteration 0” represents the matcher initialized with \texttt{RoBERTa\textsubscript{base}} and \texttt{SBERT}\texttt{(multi-qa-mpnet-base)}, thus with the correlation score equals to RoBERTa and SBERT score in Figure~\ref{fig:correlations}.}
\end{figure}

To assess the effectiveness of the iterative contrastive learning scheme, Figure~\ref{fig:curve} showcases the performance of the RoBERTa- and SBERT-based comment-response matchers on the test set across different training iterations applying iterative contrastive learning.
We can see the model's performance is improved iteratively across iterations, with the most notable enhancement occurring after the first iteration.

Even though GPT-4 achieves slightly superior correlation with humans in our experiments, from the perspective of real-world application, the cost of deploying the model is also critical. Compared with our SBERT-based matcher, prompting GPT-4 using our designed instruction template incurs an additional cost of \$4.63 on the test set based on its current pricing rate. Given the context that every year U.S. Federal Regulators receive an overwhelming volume of comment letters (usually over one million) from businesses, interest groups, and members of the public, our proposed SBERT-based matcher would be a more feasible option for such practical scenario due to its efficiency and cost-effectiveness.

\vspace{-0.7ex}
\section{Conclusion}
In this paper, we propose a simple yet effective contrastive learning approach following the iterative data construction - model updating training scheme, aiming for automatically matching the responses in policy regulations and relevant comments they respond to. Our empirical study on a real-world test set demonstrates that our proposal outperforms a set of selected benchmarks for text matching in terms of correlation with human annotations, achieves comparable performance but is more cost-effective than the most advanced gigantic large language model (i.e., GPT-4) for comments and regulator responses in larger scale.
Our proposed approach can be easily adapted to other text matching applications dealing with text in rather different complexity, such as name matching \cite{peng-etal-2015-empirical}, or extended to other more-resourced scenarios like semi-supervised settings, which we will leave as our future work.

\section{Limitations}
The main limitation of our method is that, while it provides a substantial improvement over BM25 on our task, it is not as accurate as current large language models. It seems reasonable to guess that the cost of employing GPT4 and its successors will decline over time, and at some point, the computational efficiency of our approach may not be so important. Another limitation is that our approach depends on particular aspects of our task that may not be applicable in other domains. Specifically, our unsupervised training method relies on the existence of many groups of responses and comments in the data with the property that positive pairs are only possible within a group. This lets us make good guesses about a subset of the true positive pairs with only a weak model, and generate a large number of true negative pairs by matching strings across groups. However, it is interesting to consider what other tasks and data might have a similar structure.

\section{Acknowledgement}
We thank the anonymous reviewers for their insightful comments. This research was supported by Social Sciences and Humanities Research Council of Canada (SSHRC).

\bibliography{anthology,custom}
\bibliographystyle{acl_natbib}

\clearpage
\appendix

\section{Prompt Templates for GPT-4}
\label{app:prompt_templates}
See next page.

\begin{table*}[h]
\centering
\scalebox{0.93}{
\begin{tabular}{|p{16.4cm}|}
\hline
\rowcolor{Gray}
\textbf{Content of Prompt} \\ 
\hline
\vspace{2ex} \\
I will give you a pair of comment-response texts in each turn, you should give a number between 1 and 5. The number should indicate degree of overlap between the topics discussed in the two texts and how likely it is that the agency’s response text is intended as a response to the selected comment text: \vspace{2ex}\\
\textcolor{blue}{\textbf{1 = Incorrect match.} Comment and response text are clearly discussing very different issues. The agency is definitely not responding to this comment text in the response text.} \vspace{2ex}\\
\textcolor{blue}{\textbf{2 = Poor match.} Comment and response text are somewhat related, but appear to be discussing different specific issues. It is unlikely that the agency is responding to this comment text in the response text.} \vspace{2ex}\\
\textcolor{blue}{\textbf{3 = Partial Match.} Comment and response text are discussing related issues but the degree of overlap is either imperfect or somewhat ambiguous.} \vspace{2ex}\\
\textcolor{blue}{\textbf{4 = Good match.} Comment text appears closely related to the agency’s response. It is likely that the agency is responding to this comment text.} \vspace{2ex}\\
\textcolor{blue}{\textbf{5 = Perfect match.} Comment text contains the exact argument or information that the agency is responding to in the response text. The agency is definitely responding to this specific comment text.} \vspace{3ex}\\
\textcolor{blue}{Note:} \vspace{1ex}\\
\textcolor{blue}{1. The response text could also be addressing other comments as well. This should not detract from the score. For example, if the regulator is clearly responding to two different comments A and B, and the selected comment text appears to exactly match the summary of comment A, then enter a `5'.} \vspace{2ex}\\
\textcolor{blue}{2. Sometimes there is a tension between recognizing that the comment is likely the one being discussed, and whether there is a good topic match. For example, both the comment and response might identify the commenter by name making it clear that this is the correct comment. However, if the topics do not match, the score should still be low (keep in mind this is only a sample of the comment text - it is likely that there is another omitted sample of the comment text that would be a better match).} \vspace{3ex}\\
Please give me the answer of the following comment-response pair in such format: number - explanation. \vspace{0.5ex}\\
\#\#\# \vspace{2ex}\\
Comment Text: ... \vspace{2ex}\\
Response Text: ... \vspace{2ex}\\
\hline
\end{tabular}}
\caption{The prompt templates we applied for the GPT-4 comment-response matching prediction. Text in \textcolor{blue}{blue} is the content of annotation scheme we also showed to the annotators to label our test data.}
\label{tab:your_label}
\end{table*}

\section{Details for Regulatory Data Construction}
\label{app:data_cosntruction}
Our data comes from two main sources:  Rules published in the \textit{Federal Register} from 2000-2022, downloaded in bulk XML format from govinfo.gov, and all comments submitted to \textit{regulations.gov} from 2000-2022, downloaded via the API. We extracted regulator responses to comments from the rules using a supervised classifier under development for a parallel research project. We extract comment text with the tika parser\footnote{\url{https://tika.apache.org/0.7/parser.html}}, employing OCR when necessary to extract text from image-only PDFs. The comment text is split into paragraphs, and body paragraphs are identified using a simple rule-based classifier. Finally, we group very short paragraphs (often improperly split by page breaks or other formatting issues) with adjacent paragraphs to form larger comment "chunks" 500-100 characters long. Paragraphs longer than 1000 characters are included as single chunks. Besides this rule-based chunk generation strategy, we believe topic segmentation techniques \cite{xing-etal-2020-improving, xing-carenini-2021-improving} can potentially lead to comment chunks in better quality if the training data for segmentation in reasonable size is available.

Linking comments to the appropriate rules requires additional data. We collect rule metadata from \textit{federalregister.gov} and \textit{reginfo.gov} and link \textit{regulations.gov} documents to Proposed Rules, and Proposed Rules to Rules using Federal Register document numbers, agency docket identifiers, and Regulation Identification Numbers (RIN). This gives us a database of rules where, for each rule, we can identify the set of comments that the agency would likely be responding to.

The structure of the data is important for our training strategy. Each rule may contain multiple responses, and be linked to multiple comments with several paragraphs each. We can be reasonably confident that each response in a rule is responding to a small number of paragraphs from the linked comments. It is also unlikely that that a given response is related to comment paragraphs from other rules.

When selecting the training data in our iterative algorithm, we restrict our sample to rules with 1-10 comments, and fewer than 1000 unique linked comment paragraphs. We also select at most 10 responses from each rule. This gives us a base sample of 6,727 rules, 17,452 responses, 10,456 linked comments, and 193,143 comment chunks.

To evaluate the quality of the similarity scores learned on the full training set, we used an early iteration of the model to retrieve all pairs with a score greater than 0.1 on a subset of the data. Then we grouped the pairs into bins of width 0.1 by score and kept 10 observations per bin per response. This sampling approach gives us a relatively uniform distribution of match qualities for our test sample. Finally, we sampled 4 random batches of 40 pairs from this binned sample and distributed them to human annotators. The annotators were not shown the scores used to construct the sample.

Our annotators consisted of seven students from the law program of our institution.
All of the students had been working with us for several months and were familiar with our data. The annotators were asked to score the relevance of each comment chunk to the accompanying response using a 5-point Likert scale (see Appendix~\ref{app:prompt_templates} for the annotation instructions). Each sample was assigned to multiple annotators, and we received 3-5 independent evaluations for each pair.

\end{document}